\documentclass[10pt,twocolumn,letterpaper]{article}

\usepackage{iccv}
\usepackage{times}
\usepackage{epsfig}
\usepackage{graphicx}
\usepackage{amsmath}
\usepackage{amssymb}
\usepackage{bm}
\usepackage{multirow}

\usepackage[pagebackref=true,breaklinks=true,letterpaper=true,colorlinks,bookmarks=false]{hyperref}

\iccvfinalcopy 

\def\iccvPaperID{6495} 

\ificcvfinal\fi

\begin{document}

\title{Hybrid Generative‑Discriminative Object Placement}

\author{Siyuan Zhou\\
Shanghai Jiao Tong University\\
{\tt\small ssluvble@sjtu.edu.cn}
\and
Li Niu\\
Shanghai Jiao Tong University\\
{\tt\small ustcnewly@sjtu.edu.cn}
}

\maketitle
\ificcvfinal\thispagestyle{empty}\fi

\begin{abstract}

As an important operation of image composition, object placement aims to predict the plausible placement (location, scale) for the inserted foreground object. Previous object placement methods can be divided into generative methods and discriminative methods, both of which cannot balance efficiency and effectiveness well. In this work, we propose a semi-generative method in the middle ground between them. In particular, we assign uniformly distributed anchors on the background. Then, we fuse foreground and background features to
predict the rationality score for each anchor and predict plausible placement sets for positive anchors. Extensive experiments on the OPA dataset show that our method can strike a good balance between efficiency and effectiveness.

\end{abstract}

\section{Introduction} \label{sec:intro}

Image composition, aiming to combine the foreground and background from different images as a composite image, is a crucial and popular technology for image editing, which has been extensively used in augmented reality, artistic creation, and automatic advertising \cite{WhereandWhoTan2018,MISCWeng2020,WhatWhereZhang2020}. However, the obtained composite images via simple cut-and-paste could have many issues (\emph{e.g.}, illumination, shadow, placement) that harm the image quality. 
As image composition is a complicated task involving many issues to be addressed, most previous works only focus on one of the issues, giving rise to different subtasks like image harmonization \cite{TsaiDIHarmonization2017,CongDoveNet2020,IHCISSAMCun2020}, shadow generation \cite{ARShadowGANLiu2020,hong2021shadow}, and object placement learning \cite{SyntheticTripathi2019,LearningObjPlaZhang2020,WhereandWhoTan2018}. 

\begin{figure*}[t] 
\centering 
\includegraphics[width=0.99\textwidth]{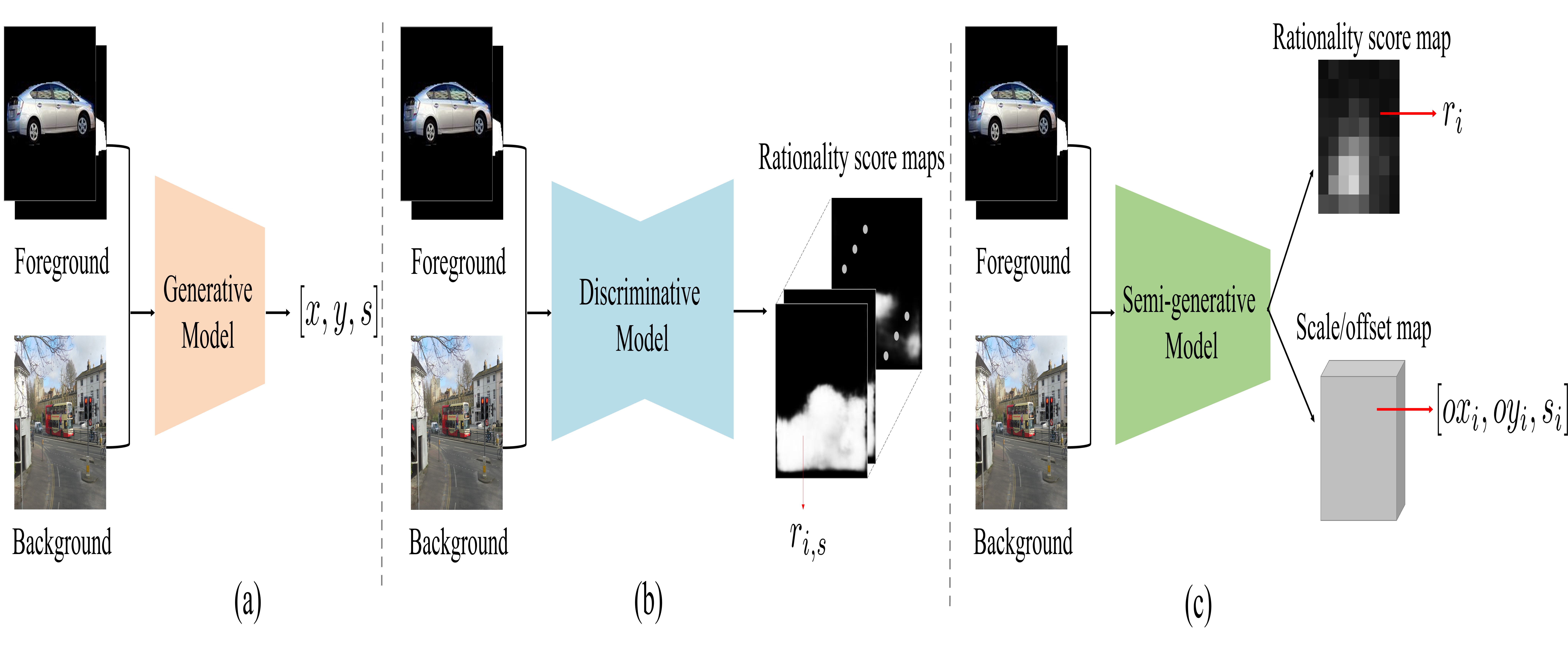}
\caption{The comparison among three paradigms for object placement learning. Given  a pair of foreground with mask and background, the generative model (a) generates one or multiple plausible placements (center location $(x,y)$ and scale $s$) for the foreground. The discriminative model (b) predicts a rationality score map for each foreground scale. $r_{i,s}$ is the rationality score for the $i$-th center location and scale $s$. Our semi-generative model (c) predicts a rationality score $r_i$ for the $i$-th anchor and also predicts multiple plausible placements (center offset $(ox_i,oy_i)$ and scale $s_i$) for each positive anchor. } 
\label{Fig:compare_paradigms} 
\end{figure*}

In this work, we focus on object placement \cite{SyntheticTripathi2019,LearningObjPlaZhang2020,WhereandWhoTan2018}, which aims to predict plausible placement (\emph{i.e.}, location, scale) of the foreground object given a pair of foreground and background. Object placement is a challenging task due to the complex factors deciding the plausibility of placement, like the depth and physical size of object, semantic context, occlusion, \emph{etc}. 
Existing object placement methods can be divided into generative methods and discriminative methods. They both take in a pair of foreground and background, and the difference lies in the output format. 
The generative methods \cite{chen2021geosim,LSCPRemez2018,LearningObjPlaZhang2020,SynthesizingGeorgakis2017} predict one or multiple plausible placements for the foreground object, as shown in Figure~\ref{Fig:compare_paradigms}(a). Discriminative methods predict a rationality score for each foreground scale and each location, which measures the plausibility of composite image obtained by placing the center of scaled foreground at this location, as shown in Figure~\ref{Fig:compare_paradigms}(b).

Nevertheless, generative methods and discriminative methods have their respective drawbacks. The generative methods often encounter mode collapse issue and the predicted placements fail to cover most reasonable locations. Besides, the prediction quality is usually lower than discriminative methods. The discriminative methods can cover all the locations with better prediction quality, but the time cost is much higher than generative methods. In this work, we attempt to fill in the middle ground between discriminative method and generative method, leading to our semi-generative method which can strike a good balance between efficiency and effectiveness.

Specifically, we first extract a foreground feature vector and a background feature map from foreground and background, respectively. Then, we pass the foreground feature vector and the background feature map through a fusion module to produce the fused feature map. We treat each location in the fused feature map as an anchor, which is associated with a set of placements with the foreground center located in its neighborhood. Each anchor is classified as either a positive anchor or a negative anchor, according to whether its associated placement set has at least one plausible placement. For each positive anchor, we attempt to predict its set of plausible placements (\emph{i.e.}, the center offset and the foreground scale). The anchor locations and the center offset lead to the center location. Then, the center location and foreground scale further lead to a complete placement. With our designed anchor mechanism and output format, we construct a network structure in the middle ground between generative method and discriminative method. 
Considering the sparse annotations of existing object placement dataset, we design a two-stage training strategy. In the first stage, we train the network using the provided sparse annotations. In the second stage, we use the prediction results to refine the annotations, and retrain our network using the updated annotations. 

Another contribution of our work lies in the dynamic fusion module to fuse foreground and background information. When we place one foreground object at one location on the background, the plausibility of this placement should depend on the contextual information surrounding this location, so we attempt to aggregate background contextual information for each location. The suitable scale of contextual information depends on the foreground information and background information at this location. For example, when placing a cat or an elephant on the ground, we need to observe the surrounding regions of different scales to check the unreasonable occlusion between this foreground object and other background objects. Therefore, we use foreground and background information to predict an offset map for deformable convolution, which aggregates different scales of background contextual information adaptively for different locations and different foreground objects. 

Following \cite{zhou2022learning}, we conduct experiments on OPA dataset \cite{liu2021opa} to verify the effectiveness of our method. 
Our major contributions can be summarized as follows. 1) We propose a semi-generative object placement method, which fills in the middle ground between discriminative methods and generative methods. 2) We design a novel dynamic fusion module to fuse foreground and background information, which can aggregate different scales of background contextual information adaptively. 3) Extensive experiments on OPA dataset demonstrate that our method strikes a good balance between efficiency and effectiveness.

\section{Related Work}

\subsection{Image Composition} \label{sec:image_composition}
Image composition targets at combining foreground and background from different source images to produce a composite image. However,  the quality of composite images obtained through simple cut-and-paste may be degraded by  the appearance, geometric, and semantic inconsistency between foreground and background. Different subtasks have been studied to address different issues in composite images. For instance, to address the inconsistent  illumination between foreground and background, image harmonization \cite{TsaiDIHarmonization2017,CongDoveNet2020,Ling2021RegionawareAI,guo2021intrinsic,IHCISSAMCun2020,sofiiuk2021foreground} adjusted the foreground illumination to match the background and produced a harmonious composite image. To address the issue of missing shadow, shadow generation methods \cite{ARShadowGANLiu2020,hong2021shadow} created plausible shadow for the composite foreground.
To address the irrational placement of composite foreground, object placement methods~\cite{MVCCPDvornik2018,WhereandWhoTan2018,WhatWhereZhang2020,ContextawareLee2018,SyntheticTripathi2019,LearningObjPlaZhang2020} predicted plausible placement (\emph{e.g.}, location, scale) for the composite foreground. In this work, we follow the research line of object placement and tend to predict reasonable location/scale for the inserted foreground.

\begin{figure*}[t] 
\centering 
\includegraphics[width=0.9\textwidth]{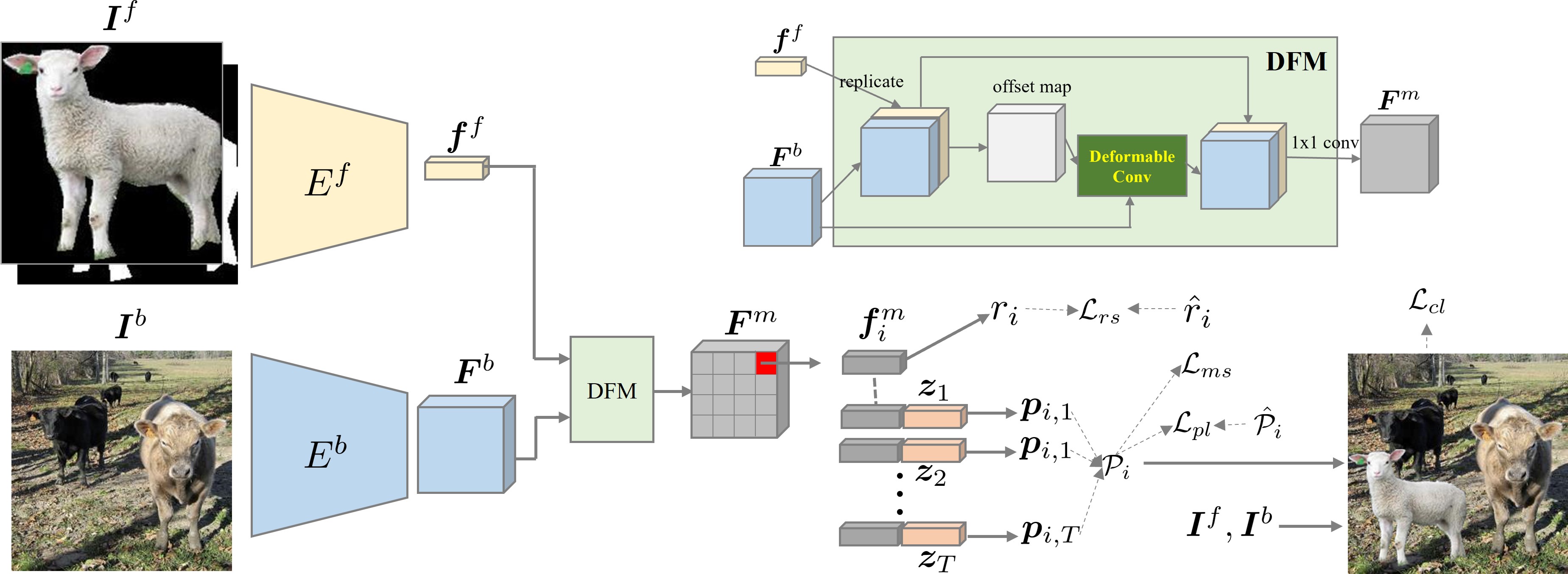}
\caption{The illustration of our network structure. We use DFM to fuse foreground feature vector $\bm{f}^f$ and background feature map $\bm{F}^b$. For each $\bm{f}_i^m$ in the fused feature map $\bm{F}^m$, we use it to predict rationality score $r_i$ and concatenate it with random vector $\bm{z}_t$ to produce a placement set $\mathcal{P}_i$.  The detailed architecture of our Dynamic Fusion Module (DFM) is shown at the top right corner. We concatenate background feature and foreground feature to produce the offset map for deformable convolution, which can dynamically aggregate multi-scale contextual information.} 
\label{Fig:network} 
\end{figure*}

\subsection{Object Placement Learning}
\label{subsection:Object_Placement_Learning}
As an inevitable step in image composition, object placement is helpful for many application scenarios like artistic design and automatic advertising~\cite{WhatWhereZhang2020}.
Existing methods on object placement learning can be roughly divided into generative methods and discriminative methods. Given a pair of foreground and background, the generative methods aim to generate one or more plausible placements for the foreground. For example, the methods \cite{STGANChen2018,zhan2019spatial,SyntheticTripathi2019,azadi2020compositional} predicted one plausible placement for the foreground. \cite{LearningObjPlaZhang2020,zhou2022learning} injected a random vector into the network and predicted multiple plausible placements by sampling the random vector. 
Generative methods are efficient in generating a few plausible placements, but the quality and diversity are lower than discriminative methods. 
Early discriminative method \cite{liu2021opa} predicted whether the foreground placement in a composite image is plausible, which is also dubbed as object placement assessment \cite{liu2021opa}.  Then, \cite{niu2022fast}  proposed a fast object placement assessment method to accelerate \cite{liu2021opa}, which takes in a scaled foreground and a background to predict all reasonable locations for this scaled foreground. Nonetheless, the accelerated version \cite{niu2022fast} is still much slower than generative methods.

In this work, we attempt to fill the middle ground between generative methods and discriminative methods. We proposed a semi-generative model which achieves comparable performances with discriminative methods at much lower cost.

\section{Our Method}

Suppose that we have a background image $\bm{I}^b\in\mathcal{R}^{H\times W \times 3}$ and a foreground image $\bm{I}^f\in\mathcal{R}^{H\times W \times 3}$ with its binary object mask $\bm{M}^f\in\mathcal{R}^{H\times W}$ delineating the foreground object, where $H$ and $W$ are image height and width respectively. The goal of object placement is predicting plausible placements (\emph{e.g.}, scale, location) for the foreground object. As claimed in Section~\ref{sec:intro}, the output formats of generative methods \cite{LearningObjPlaZhang2020,zhou2022learning} and discriminative methods \cite{niu2022fast} differ a lot. 
In this work, we design a semi-generative approach whose output format is a mixture of generative methods and discriminative methods. 

\subsection{Overall Network Structure}
\begin{figure}[t] 
\centering 
\includegraphics[width=0.47\textwidth]{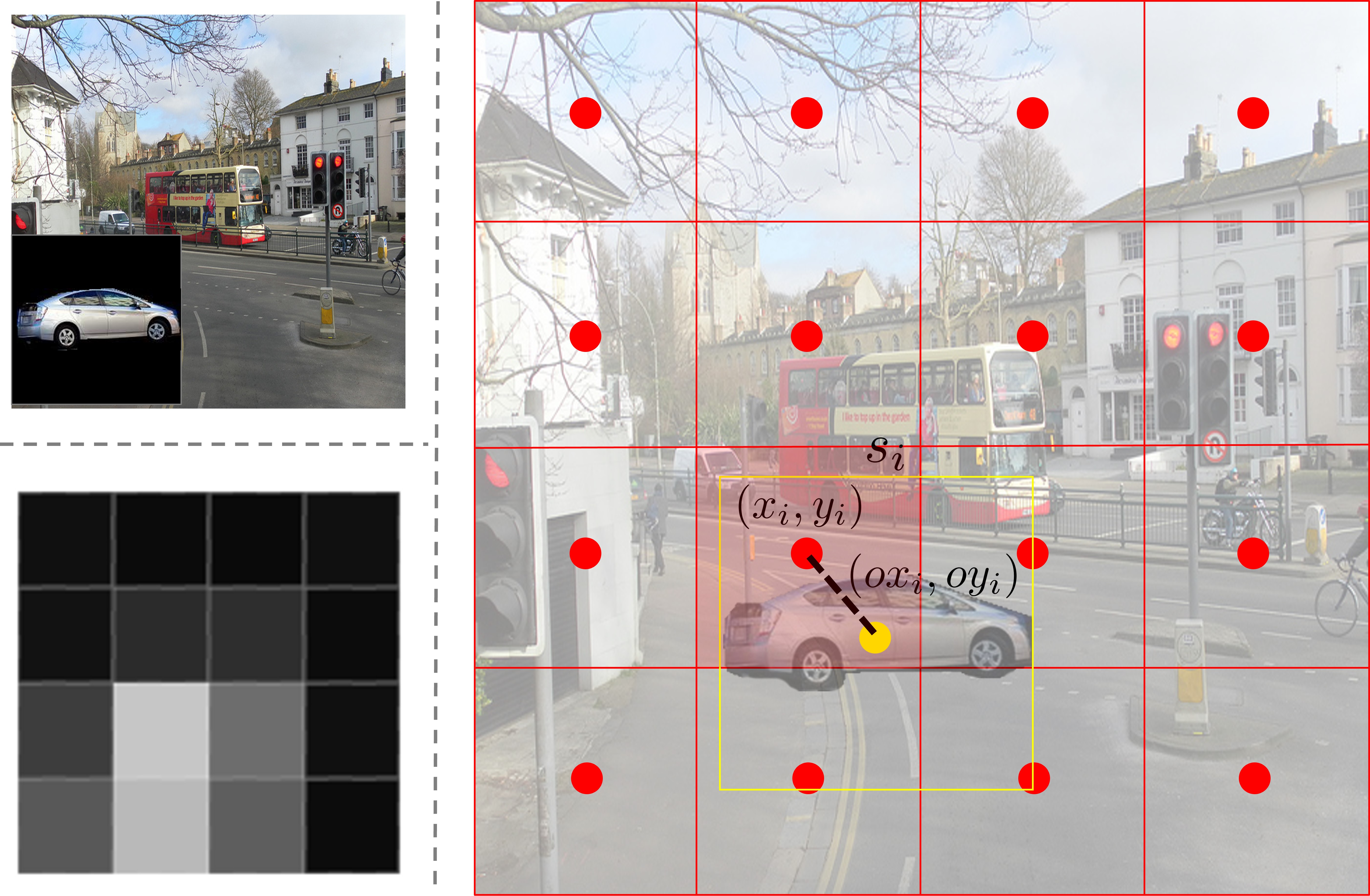}
\caption{We suppose that an image is divided into $4 \times 4$ cells, corresponding to $16$ anchors (red dot). In the left column, we show the foreground/background and the rationality score map. In the right column, we show one foreground placement, which is associated with the $i$-th anchor because the foreground center is within its neighborhood (red cell). The location of foreground center can be calculated based on the anchor location $(x_i, y_i)$ and center offset $(ox_i, oy_i)$. The foreground placement (yellow bounding box) is represented by the foreground center location $(x_i\!+\!ox_i, y_i\!+\!oy_i)$ and the foreground scale $s_i$.} 
\label{Fig:anchor} 
\end{figure}

We use foreground encoder $E^f$ and background encoder $E^b$ to extract a foreground feature map and a background feature map, respectively. Following \cite{zhou2022learning}, $E^f$ and $E^b$ share the same backbone network with different heads. 
$E^f$ takes in the concatenation of foreground image $\bm{I}^f$ and its binary object mask $\bm{M}^f$, producing the foreground feature map. We perform global average pooling over the foreground feature map and get the foreground feature vector $\bm{f}^f \in \mathcal{R}^{c}$. 
$E^b$ takes in the background $\bm{I}^b$ and produces the background feature map $\bm{F}^b\in\mathcal{R}^{h\times w \times c}$. We treat each pixel in the background feature map as an anchor, leading to $n=h\times w$ anchors. As shown in Figure~\ref{Fig:anchor}, the location of the $i$-th anchor on the background image is assumed to be the center of the $i$-th cell, after uniformly dividing the background image into $h\times w$ cells. The $i$-th cell is also called the neighborhood of the $i$-th anchor.
The rationality score $r_i$ of the $i$-th anchor indicates whether exists plausible placement with the foreground center in the neighborhood of the $i$-th anchor. Based on the rationality scores, we categorize all $n$ anchors into positive anchors and negative anchors. For positive anchors, we attempt to predict their plausible placement sets, in which one placement includes the center offset and foreground scale. On the premise of anchor location, center offset, and foreground scale, we can obtain a plausible composite image.

\subsection{Dynamic Fusion Module}

To predict the rationality score and placement set for each anchor, we integrate the foreground feature vector $\bm{f}^f$ with each pixel-wise  feature vector in the background feature map  $\bm{F}^b$. One simple way is spatially replicating $\bm{f}^f$ and concatenating them with $\bm{F}^b$ channel-wisely. However, when a foreground object is placed at certain location on the background, the plausibility of this placement relies on the nearby contextual information, so we need to aggregate background contextual information for each location. The suitable scale of contextual information to be aggregated depends on the foreground and background information at this location. 
To this end, we design a Dynamic Fusion Module (DFM) to fuse $\bm{f}^f$ and $\bm{F}^b$, which automatically decides the scale of contextual information for each location based on the foreground and background information.

As shown in Figure~\ref{Fig:network}, we spatially replicate $\bm{f}^f$ to get $\bm{F}^f$, which is concatenated with $\bm{F}^b$. The concatenation passes through one conv layer to produce an offset map for deformable convolution \cite{dai2017deformable}. The entry at location $\bm{l}$ in the offset map contains the spatial offsets $\{\Delta \bm{o}_{\bm{l},k}|_{k=1}^9\}$ for a $3\times 3$ convolution kernel centered at this location. Based on the offset map, we perform deformable convolution \cite{dai2017deformable} on $\bm{F}^b$. Formally, assuming that the kernel weights are $\{\bm{w}_k|_{k=1}^9\}$ and the relative locations in the kernel are $\bm{o}_k\in\{(-1,-1),(-1,0),\ldots,(1,1)\}$, the deformable convolution operation on $\bm{F}^b$ with kernel center at the location $\bm{l}$ can be represented by
\begin{eqnarray}
y(\bm{l}) = \sum_{k=1}^9 \bm{w}_k\cdot \bm{F}^b(\bm{l}+\bm{o}_k+\Delta \bm{o}_{\bm{l},k}),
\end{eqnarray}
in which $y(\bm{l})$ is the output value. For more details of deformable convolution, please refer to \cite{dai2017deformable}.

In this way, we aggregate different scales of background contextual information adaptively for different locations. The output of deformable convolution is concatenated with $\bm{F}^f$. The concatenation further passes through a $3\times 3$ conv layer to produce the fused feature map $\bm{F}^m$. 

\subsection{Rationality and Placement Prediction}

We denote the $i$-th feature vector in the fused feature map $\bm{F}^m$ as $\bm{f}^m_i$, which is used to predict the rationality score and a set of plausible placements for the $i$-th anchor. 
The rationality score $r_i$ within the range of $[0,1]$ indicates the existence of plausible placement centered around this location (the foreground center is in the neighborhood of this anchor). Assuming that we have the ground-truth rationality score $\hat{r}_i$, the predicted
$r_i$ is supervised using binary cross-entropy loss:
\begin{eqnarray}
\mathcal{L}_{rs} = -\hat{r}_i \log r_i - (1-\hat{r}_i) \log (1-r_i).
\end{eqnarray}

We refer to the anchors with rationality scores larger than the threshold $0.5$ as positive anchors and the other anchors as negative anchors. We only supervise the predicted placements of positive anchors, while leaving those of negative anchors unattended. 
To predict multiple plausible placements for each positive anchor $i$, we concatenate  $\bm{f}^{m}_i$ with a random vector $\bm{z}\!\sim\! \mathcal{N}(0,1)$. The concatenated vector is sent into a prediction head to produce one placement $\bm{p}_i= (ox_{i}, oy_{i}, s_{i})$, in which $(ox_i,oy_i)$ is the center offset and $s_i$ is the foreground scale. By denoting the location of the $i$-th anchor as $(x_i, y_i)$, the predicted foreground object center is  $(x_i+ox_i, y_i+oy_i)$, which should be constrained within the neighborhood of the $i$-th anchor. Note that we have a common scale $s_i$ for both height and width of the foreground object, because we do not change the aspect ratio of foreground object.  

By sampling $\bm{z}$ multiple times, we can get multiple placements for each anchor. Specifically, when sampling $T$ times, we have $\{\bm{z}_1,\ldots, \bm{z}_T\}$ and produce a placement set $\mathcal{P}_i=\{\bm{p}_{i,1},\ldots, \bm{p}_{i,T}\}$ for the $i$-th anchor, in which $\bm{p}_{i,t} = (ox_{i,t}, oy_{i,t}, s_{i,t})$. 
To ensure the diversity of predicted placements, we employ the mode seeking loss \cite{mao2019mode} to enforce the predictions with different $\bm{z}$ to be different:
\begin{eqnarray}  \label{eqn:mode_seeking_loss}
\!\!\!\!\!\!\!\!&&\mathcal{L}_{ms} =   \sum_{t\neq t'}   \frac {|| \bm{z}_{t} - \bm{z}_{t'}||_1} {||\bm{p}_{i,t} - \bm{p}_{i,t'} ||_1}.
\end{eqnarray}
Intuitively, when $|| \bm{z}_t - \bm{z}_{t'}||_1$ is large, we expect $\bm{p}_{i,t}$ and $\bm{p}_{i,t'}$ to be considerably different, which can push the prediction head to search more modes to produce diverse placements. 
Let us assume that we have a set of ground-truth plausible placements for the $i$-th anchor
$\hat{\mathcal{P}}_i=\{\hat{\bm{p}}_{i,1},\ldots,\hat{\bm{p}}_{i,K_i}\}$, in which $K_i$ is the number of ground-truth plausible placements for the $i$-th anchor. Note that we set $T\geqslant  K_i,\,\forall i$.
When $K_i< T$, we pad  $\hat{\mathcal{P}}_i$  to match the size $T$.
We need to measure the distance between $\mathcal{P}_i$ and  $\hat{\mathcal{P}}_i$, which involves matching two placement sets. 
Inspired by \cite{carion2020end}, we perform bipartite matching between $\mathcal{P}_i$ and $\hat{\mathcal{P}}_i$, seeking for an index mapping $\hat{\sigma}$ such that the matching cost is minimized. Formally,
\begin{eqnarray}
\hat{\sigma} = \arg\min_{\sigma} \sum_{t=1}^T L_2(\hat{\bm{p}}_{i,t}, \bm{p}_{i,\sigma(t)}),
\end{eqnarray}
in which $L_2(\hat{\bm{p}}_{i,t}, \bm{p}_{i,\sigma(t)})$ is the $L_2$ distance between $\hat{\bm{p}}_{i,t}$ and $\bm{p}_{i,\sigma(t)}$ when neither $\hat{\bm{p}}_{i,t}$ nor $\bm{p}_{i,\sigma(t)}$ is $\emptyset$. Otherwise,  $L_2(\hat{\bm{p}}_{i,t}, \bm{p}_{i,\sigma(t)})=0$. Then, the matching loss between $\mathcal{P}_i$ and  $\hat{\mathcal{P}}_i$ can be calculated as 
\begin{eqnarray}
\mathcal{L}_{pl} = \sum_{t=1}^T \mathcal{L}_2(\hat{\bm{p}}_{i,t}, \bm{p}_{i,\hat{\sigma}(t)}).
\end{eqnarray}
For more details of matching loss, please refer to \cite{carion2020end}. 

In addition, we employ the slow object placement assessment (SOPA) model \cite{liu2021opa} to supervise the predicted placements of positive anchors. The SOPA model, denoted by $D$, is a binary classification model to predict the score of a composite image ($1$ for positive and $0$ for negative). Given the predicted placements of positive anchors, we can obtain the composite images $\bm{I}^c$ and send them to $D$ to maximize the predicted classification scores. The rationality classification loss is given by
\begin{eqnarray}
\mathcal{L}_{cl} = -\log D(\bm{I}^c). 
\end{eqnarray}

So far, the collection of all training losses is
\begin{eqnarray} \label{eqn:total_loss}
\mathcal{L}_{all} = \lambda_1\mathcal{L}_{rs} +  \mathcal{L}_{ms} +\lambda_2 \mathcal{L}_{pl}  +  \mathcal{L}_{cl},
\end{eqnarray}
in which $\lambda_1$ and $\lambda_2$ are trade-off parameters. 

\subsection{Training Strategies}

Following \cite{zhou2022learning}, we use the object placement assessment (OPA) dataset contributed by \cite{liu2021opa}, which provides sparse annotations for partial positive and negative placements (location, scale). We assign annotated placements to their corresponding anchors and refer to the anchors associated with at least one annotated placement as annotated anchors. The unannotated anchors are ignored. 
If one annotated anchor has at least one positive placement, then this annotated anchor is a positive anchor ($\hat{r}_i=1$) and we collect its positive placement set $\hat{\mathcal{P}}_i$. Otherwise, this annotated anchor is a negative anchor ($\hat{r}_i=0$).  We use the supervision information of annotated anchors to train the model. However, the above supervision information is very noisy. In particular, the negative anchors may also have unannotated positive placements, which should actually be positive anchors. Besides, the positive anchors could have many potential positive placements beyond the annotated ones. Therefore, we add a self-training stage to retrain the model using updated supervision information.

In the self-training stage, we update the annotations as follows. The original positive anchors are still positive. For the other anchors, if their predicted rationality scores $r_i$ are larger than a threshold $\tau_c$, we set them as positive anchors. All the other anchors are set as negative anchors. For the positive anchors, we augment their positive placement sets
$\hat{\mathcal{P}}_i$. Specifically, for each positive anchor, we randomly sample $50$ random vectors $\bm{z}$ and collect the placements whose classification scores predicted by $D$ are larger than a threshold $\tau_s$. We add these placements to $\hat{\mathcal{P}}_i$. If the size of augmented $\hat{\mathcal{P}}_i$ exceeds $T$, we adopt the top $T$ placement with the highest classification scores. Finally, we retrain our model using the updated annotations.
\begin{table*}[t]
  \centering
  \caption{Quantitative comparison of different methods on OPA dataset.}
  \label{table:sota}
  \begin{tabular}{c|c|c|c|c|c|c}
	\hline
	\multirow{2}*{Method} & \multicolumn{3}{c|}{\emph{Plausibility}} & \multicolumn{1}{c|}{\emph{Diversity}} & \multicolumn{2}{c}{\emph{Complexity}}  \\ \cline{2-7} 
	& User Study$\uparrow$ & Acc.$\uparrow$ & FID$\downarrow$ & LPIPS$\uparrow$ & FLOPs(G) & Time(s)\\
	\hline
	TERSE~\cite{SyntheticTripathi2019} & 0.104 & 0.679 & 46.94 & 0 & 37.25 & 0.040 \\
	PlaceNet~\cite{LearningObjPlaZhang2020} & 0.109 & 0.683 & 36.69 & 0.160 & 4.76 & 0.039\\
	GracoNet~\cite{zhou2022learning} & 0.183 & 0.847 & 27.75 & 0.206 & 43.98 & 0.044 \\
	\hline
	FOPA~\cite{niu2022fast} & 0.303 & 0.896 & 25.91 & 0.216 & 321.05 & 0.258 \\
	\hline
	Ours & 0.301 & 0.890 & 26.02 & 0.218 & 39.49 & 0.041 \\
	\hline
  \end{tabular}
\end{table*}


\section{Experiments}

\subsection{Dataset} \label{sec:dataset}
Following \cite{zhou2022learning}, we conduct experiments on OPA dataset \cite{liu2021opa}, which contains composite images with annotated binary rationality labels. In detail, OPA dataset contains $62,074$ (\emph{resp.}, $11,396$) annotated composite images for training (\emph{resp.}, testing). The training (\emph{resp.}, test) set consists of $21,376$ (\emph{resp.}, $3,588$) positive and  $40,698$ (\emph{resp.}, $7,808$) negative samples. 
There are $4,137$ different foreground objects from $47$ categories and $1,389$ different background images in the dataset. The foregrounds/backgrounds in the training set and test set have no overlap. 

To enable our model to be trained on the training set, we first transform the annotation format of the training set. 
Specifically, given a pair of foreground and background, we collect the annotated composite images (positive or negative composite images) with the given foreground/background, and obtain the
annotated placements (positive or negative placements), because one composite image corresponds to one foreground placement. 
If an anchor is associated with annotated placements, that is, there exists annotated placement whose foreground center is in the neighborhood of this anchor, we refer to this anchor as annotated anchor.
If an annotated anchor has at least one positive placement, this anchor is a positive anchor and we collect its associated positive placements as the ground-truth placement set. Otherwise, this annotated anchor is a negative anchor. After acquiring the positive/negative anchors and the ground-truth  placement sets of positive anchors, we train our model on the transformed training set. 

\subsection{Evaluation Metrics} \label{sec:evaluation_metrics}

For model evaluation, we strictly follow the setting in \cite{zhou2022learning}. In particular, our model takes one test pair (foreground-background pair of each positive composite image in the test set) as input, and generates composite images.
Following \cite{zhou2022learning}, we adopt user study, accuracy, FID~\cite{FIDHeusel2017} to evaluate the plausibility of generated composite images, and adopt LPIPS~\cite{zhang2018unreasonable} to evaluate the diversity of generated composite images. 

For plausibility evaluation, our model takes in one test pair and generates a composite image. Specifically, we choose the positive anchor with the highest rationality score and randomly sample one $\bm{z}$ to generate a composite image. 
We use the binary classifier provided by \cite{zhou2022learning} to predict the rationality of generated composite images. \emph{Accuracy} is defined as the proportion of generated composite images that are classified as positive by the binary classifier. \emph{FID} is calculated between generated composite images and positive composite images in the test set. \emph{User study} shows the proportion that each method is selected as the best one (see details in Supplementary).

For diversity evaluation,  our model takes in a test pair and generates $10$ composite images by randomly selecting $10$ positive anchors and randomly sample one $\bm{z}$ for each positive anchor. For each test pair, we compute the average pairwise \emph{LPIPS} of $10$ generated composite images. After that, we calculate the average of \emph{LPIPS} over all test pairs. 

\subsection{Implementation Details}
The foreground encoder $E^f$ and the background encoder $E^b$ have the same backbone network and two separate heads. The shared backbone contains the beginning 34 layers (including and before the fourth $\mathrm{MaxPool}$ layer) of VGG16-BN~\cite{simonyan2015deep} pretrained on ImageNet~\cite{deng2009imagenet}. Both the foreground head and the background head contain three groups of ($3\times 3$ $\mathrm{Conv}$, $\mathrm{BN}$, $\mathrm{ReLU}$). The foreground head is ended with another $\mathrm{GlobalAvgPool}$ to extract the foreground feature vector, while the background head is ended with another $h\times w$ $\mathrm{AdaptiveAvgPool}$ to extract the background feature map. Images are resized to $256\times 256$ and normalized before we fed them into the network. We train the model with batch size 16 for 20 epochs (including the self-training stage) on a single RTX 3090 GPU. We adopt Adam optimizer with the initial learning rates being $2 \times 10^{-5}$ for backbone and $2 \times 10^{-4}$ for the remaining parts. 
For hyper-parameters, we set the number of anchors as $n=h\times w=8\times 8=64$, the sampling times for each anchor as $T=10$. In Eqn.~\ref{eqn:total_loss}, we set $\lambda_1=10, \lambda_2=5$. In the self-training stage, we set $\tau_c=0.6, \tau_s=0.85$. 

\begin{figure*}[t] 
\centering 
\includegraphics[width=0.93\textwidth]{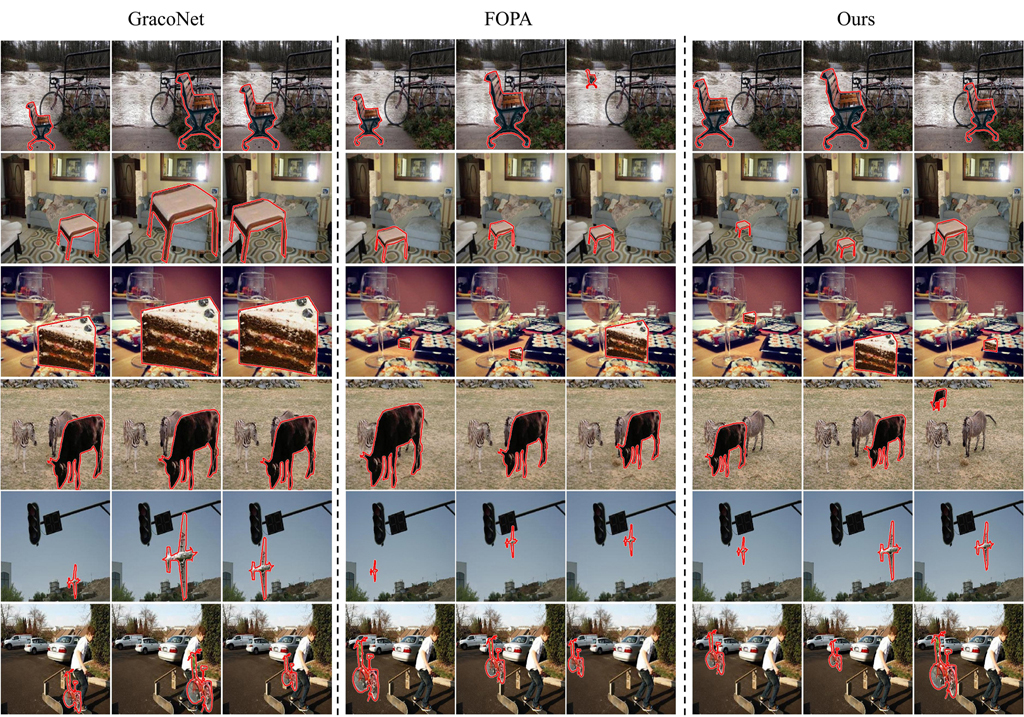}
\caption{For each foreground-background pair, we show three composite images generated by GracoNet~\cite{zhou2022learning}, FOPA~\cite{niu2022fast}, and our method.} 
\label{Fig:cmp_baselines_main} 
\end{figure*}

\subsection{Comparison with Baselines} \label{sec:baselines}
We compare with two groups of baselines: generative object placement methods and discriminative object placement methods. For generative methods, we compare with TERSE~\cite{SyntheticTripathi2019}, PlaceNet~\cite{LearningObjPlaZhang2020}, and GracoNet \cite{zhou2022learning}. 
They take in a pair of foreground and background, generating one or multiple composite images, which is similar to our method. 
When evaluating diversity (\emph{resp.}, plausibility), \cite{SyntheticTripathi2019,zhou2022learning} produce $10$ (\emph{resp.}, $1$) composite images by sampling the random vector for $10$ (\emph{resp.}, $1$) times. 
Note that TERSE~\cite{SyntheticTripathi2019} does not have random vector and could only produce one composite image, so its LPIPS is 0. 

For discriminative method, we compare with FOPA~\cite{niu2022fast}, which requires a background and a scaled foreground as input to produce a rationality score map. 
As described in \cite{niu2022fast}, we predict $16$ rationality score maps for $16$ discrete foreground scales, and obtain one composite image with the highest pixel-wise rationality score across $16$ maps. When evaluating the diversity, we randomly choose $10$ composite images whose rationality scores exceed the threshold $0.5$. 

We also compare the efficiency between different methods. We report FLOPS and inference time taken to process a foreground-background pair. We test the inference time on a single RTX 3090 GPU with PyTorch-1.7~\cite{paszke2019pytorch}, by running over all test pairs and calculating the average.

The results of different methods are summarized in Table \ref{table:sota}. 
For plausibility evaluation, \emph{accuracy} is more reliable than \emph{FID}. FOPA outperforms generative methods on accuracy by a large margin. For diversity evaluation, LPIPS of FOPA does not surpass GracoNet significantly. One possible reason is that GracoNet tends to generate large composite foregrounds (see Figure~\ref{Fig:cmp_baselines_main}), and the relocation of large composite foreground leads to high LPIPS. Moreover, merely persuing high LPIPS is not meaningful because randomly placing the composite foreground could also achieve high LPIPS. 
For efficiency, the computational cost of FOPA is significantly larger than generative methods, due to its complex network structure.

Based on Table \ref{table:sota}, our method outperforms generative methods significantly and achieves comparable results with FOPA. In terms of efficiency, our method is comparable with GracoNet and requires much less computational cost than FOPA. Therefore, our method strikes a good balance between efficiency and effectiveness.

\subsection{Ablation Studies}

\begin{table}[t]
  \centering
  \caption{Quantitative comparison of different ablated versions of our method on OPA dataset.}
  \label{table:ablate}
  \begin{tabular}{c|c|c|c|c}
	\hline
	\multirow{2}*{Row} & \multirow{2}*{Method} & \multicolumn{2}{c|}{\emph{Plausibility}} & \multicolumn{1}{c}{\emph{Diversity}} \\ \cline{3-5} 
	& & Acc.$\uparrow$ & FID$\downarrow$ & LPIPS$\uparrow$ \\
	\hline
	1 & w/o DC & 0.871 & 26.85 & 0.212\\
    2 & w/o $\mathcal{L}_{ms}$ & 0.845 & 27.67 & 0.176 \\
    3 & w/o $\mathcal{L}_{pl}$ & 0.832 & 28.83 & 0.196 \\
    4 & w/o $\mathcal{L}_{cl}$ & 0.712 & 44.86 & 0.188 \\
    5 & $1$st stage & 0.869 & 26.93 & 0.211 \\
	\hline
	6 & Ours & 0.890 & 26.02 & 0.218\\
	\hline
  \end{tabular}
\end{table}

We evaluate the effectiveness of each component and loss term in Table \ref{table:ablate}. First, we replace deformable convolution in our FAD module with vanilla $3\times 3$ convolution, leading to the results in row 1. The performance becomes worse, which demonstrates that it is helpful to dynamically aggregate contextual information. Then, we remove the mode seeking loss $\mathcal{L}_{ms}$ (row 2), after which the diversity is significantly degraded. Note that the predicted placements of positive anchors are supervised by the matching loss $\mathcal{L}_{pl}$ and the classification loss $\mathcal{L}_{cl}$. We remove either of them (row 3-4), which shows that both loss terms contribute to the performance. $\mathcal{L}_{cl}$ plays a more important role than $\mathcal{L}_{pl}$, because it pushes all generated composites images from positive anchors to be positive. 

Recall that we take a two-stage training strategy. We also report the performance in the first stage without self-training. By comparing the performances between two stages (row 5 \emph{v.s.} row 6), we conclude that the self-training stage could boost the performance by refining the annotations.

\begin{figure}[t] 
\centering 
\includegraphics[width=0.45\textwidth]{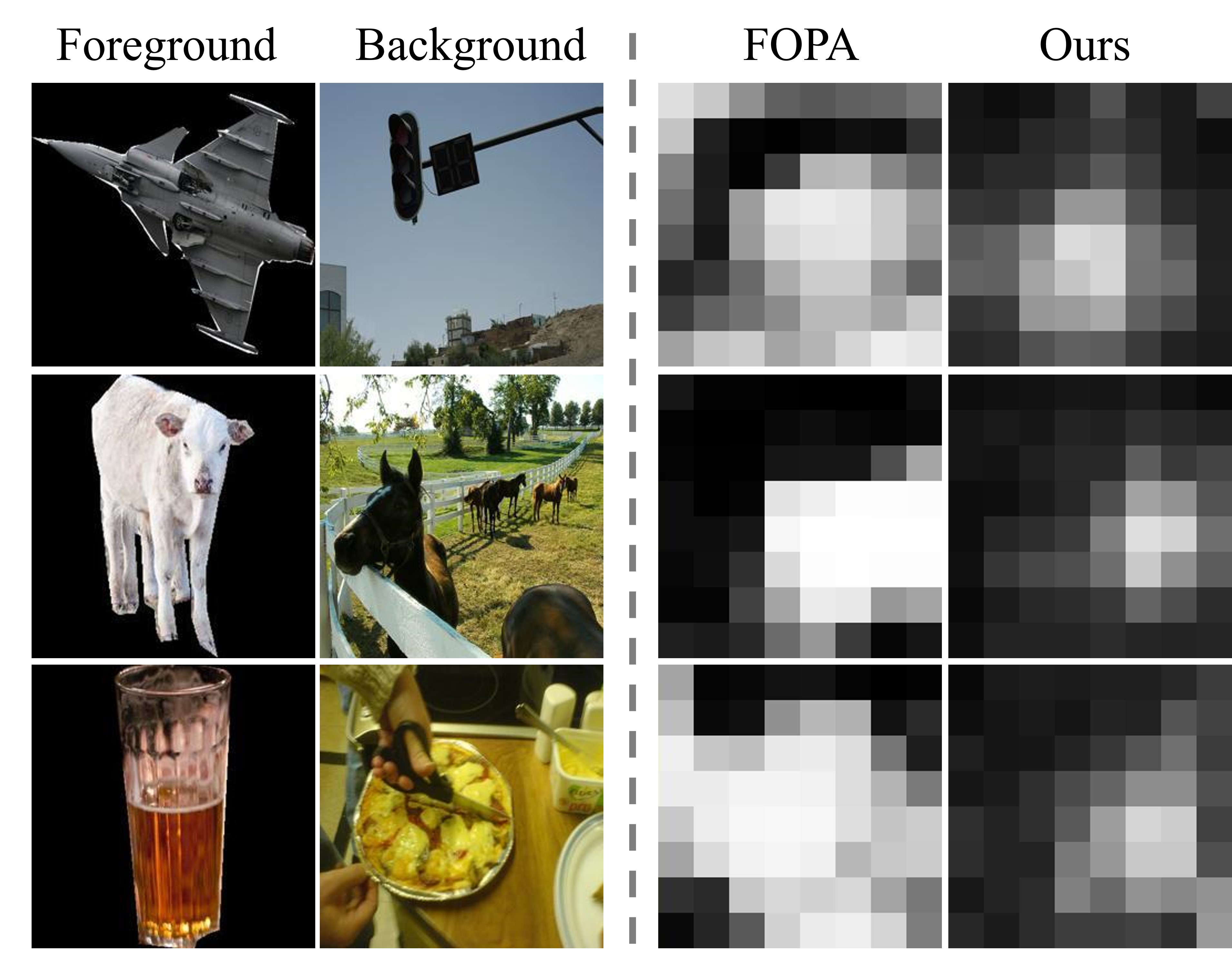}
\caption{The comparison of rationality score map between FOPA~\cite{niu2022fast} and our method. From left to right, we show the foreground, background, the rationality score map generated by FOPA and our method. } 
\label{Fig:heatmap_comparison} 
\end{figure}

\begin{figure}[t] 
\centering 
\includegraphics[width=0.45\textwidth]{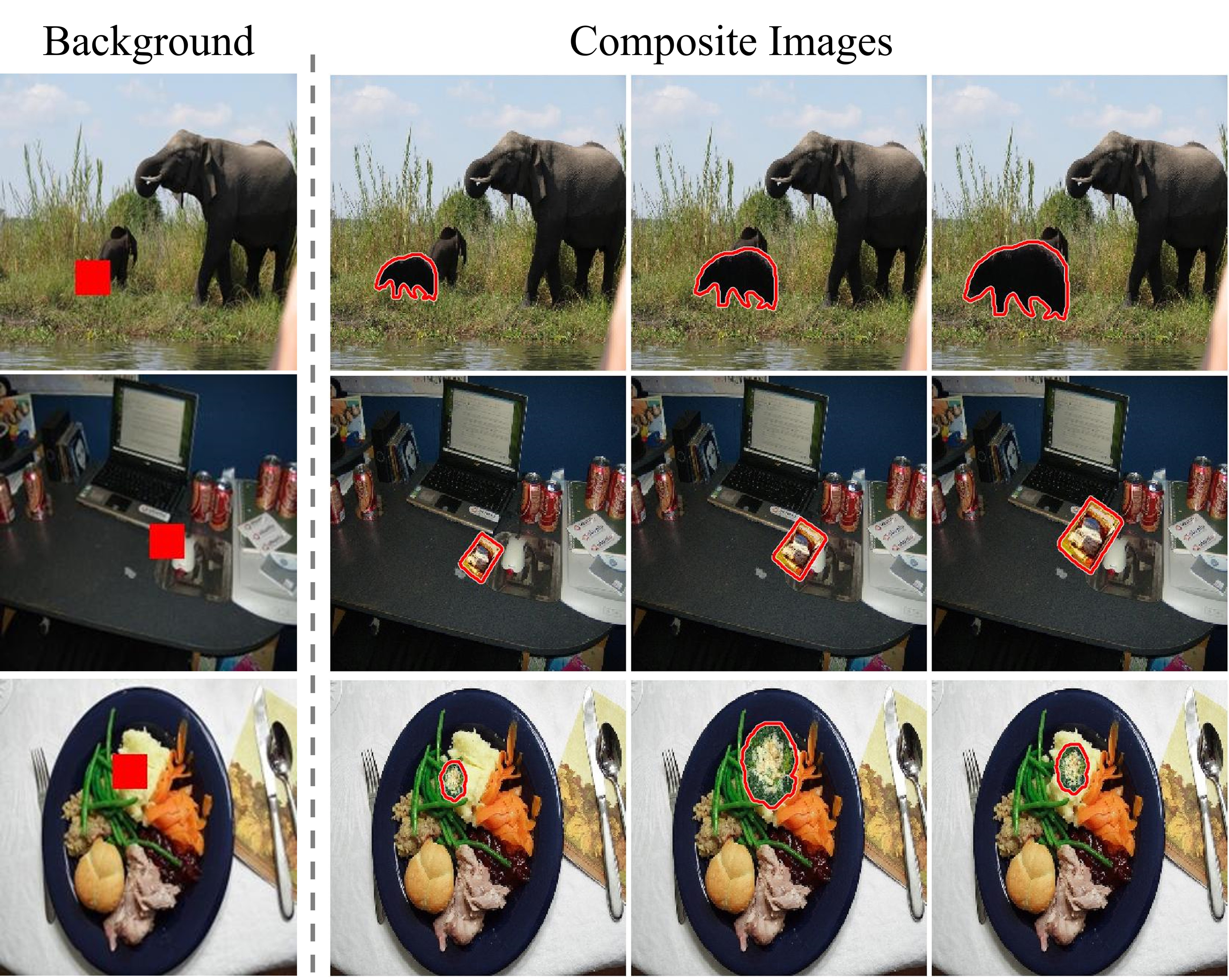}
\caption{For each pair of foreground and background, we show three generated composite images for the same positive anchor with the highest rationality score. The neighborhood of each positive anchor is marked with red box on the background.}
\label{Fig:same_anchor} 
\end{figure}

\subsection{Visualization Results}

\noindent\textbf{Generated composite images:} We show the composite images generated by different methods in Figure \ref{Fig:cmp_baselines_main}. We compare with the competitive generative baseline GracoNet (see Table~\ref{table:sota}) and the discriminative baseline FOPA. 
Given a pair of foreground and background, we show three composite images generated by two baselines and our method. The details of generating multiple placements have been introduced in Section \ref{sec:evaluation_metrics} for our method and Section \ref{sec:baselines} for \cite{zhou2022learning,niu2022fast}. 

We observe that GracoNet tends to generate large and unreasonable composite foreground (row 2, 3). The location diversity is limited compared with the other two methods. For example, GracoNet places the sheep on the right (row 4) and the bicycle surrounding the person (row 6), despite the fact that there are many other reasonable locations. 
Compared with GracoNet, the generated composite images of FOPA have higher quality and cover more reasonable locations. However, there still exist unreasonable occlusions (\emph{e.g.}, airplane occludes the traffic light in row 5) and unreasonable locations (\emph{e.g.}, chair in the water in row 1). Our method achieves comparable or better results than FOPA, and has the ability to generate diverse and plausible placements. Moreover, our method is significantly faster than FOPA (see Section \ref{sec:baselines}). Thus, our method can balance effectiveness and efficiency better than previous generative and discriminative methods.  

\noindent\textbf{Rationality score map:} In Figure~\ref{Fig:heatmap_comparison}, we show the rationality score map indicating positive anchors and negative anchors. Recall that positive anchor means that there exists at least one positive placement with foreground center in its neighborhood. Based on the rationality score map, our method can roughly tell the positive anchors from negative anchors. 

We also compare with the rationality score map predicted by FOPA \cite{niu2022fast}. Note that the meanings of rationality score map of our method and \cite{niu2022fast} are different. \cite{niu2022fast} predicts one rationality score map for each foreground scale, in which each entry represents the rationality of composite image obtained by moving the center of scaled foreground to this location. For meaningful comparison, we first generate  $16$  rationality score maps with size $H\times W$ for $16$ discrete foreground scales and concatenate them channel-wisely, resulting in a rationality score tensor. Then, we perform max pooling within each $\frac{H}{h}\times \frac{W}{w}\times 16$ volume and get the pooled rationality score map with size $h\times w$, because the maximum value within each volume roughly indicates the existence of positive placement for this anchor. We compare the rationality score maps between FOPA and our method in Figure~\ref{Fig:heatmap_comparison}. It can be seen that the positive anchors (\emph{e.g.}, sky for the airplane in row 1, unoccupied  grassland for the sheep in row 2) identified by two methods basically coincide with each other. In some cases, our method can predict better rationality score map. For example, in row 3, our method identifies the free space on the table while FOPA includes extra unreasonable locations. 
The above visualization results demonstrate the effectiveness of our predicted rationality score maps.

\noindent\textbf{Diversity for one anchor:} To better demonstrate the diversity of our generated composite images, we show multiple results for the same positive anchor in Figure \ref{Fig:same_anchor}. Given a pair of foreground and background, we find the positive anchor with the largest rationality score. Then, we randomly sample $\bm{z}$ three times for this positive anchor and generate three composite images. It can be seen that the foreground center lies in the neighborhood of positive anchor. Under this constraint, the generated composite images still exhibit diversity in terms of scale and location. 

\noindent\textbf{Offsets in FAD module:} We visualize the learnt offsets in our FAD module. We observe that the learnt offsets vary dynamically according to the foreground information and background context. The detailed results are left to Supplementary due to the space limitation. 

\subsection{Hyper-parameter Analyses and Failure Cases}

We investigate the impact of hyper-parameters in our method and show several failure cases of our method, which can be found in Supplementary. 

\section{Conclusion}

In this work, we have developed an anchor-based semi-generative object placement learning method, aiming to combine the advantages of generative methods and discriminative methods. Given a pair of foreground and background, our method can produce high-quality rationality score map as well as diverse and plausible placement sets. Comprehensive experiments on OPA dataset have demonstrated the superiority of our proposed method.

{\small
\bibliographystyle{ieee_fullname}
\bibliography{egbib}
}

\end{document}


\title{Supplementary for Hybrid Generative‑Discriminative Object Placement}

\author{Siyuan Zhou\\
Shanghai Jiao Tong University\\
{\tt\small ssluvble@sjtu.edu.cn}
\and
Li Niu\\
Shanghai Jiao Tong University\\
{\tt\small ustcnewly@sjtu.edu.cn}
}

\maketitle
\ificcvfinal\thispagestyle{empty}\fi

In this document, we provide additional materials to supplement the main paper. In Section \ref{sec:hyperparameter}, we investigate the impact of hyper-parameters. In Section~\ref{sec:user_study}, we provide the details for user study. 
In Section \ref{sec:visual_comparison}, we provide more visualization results compared with competitive baselines. In Section \ref{sec:offset_visualization}, we visualize the offsets predicted by our dynamic fusion module. In Section \ref{sec:our_visualization}, we provide more visualization results of our method. In Section \ref{sec:failure_cases}, we discuss the limitation of our method. 

\section{Hyper-parameter Analyses}\label{sec:hyperparameter}

In this section, we investigate the impact of hyper-parameters used in our method. 

\begin{figure}[t]
\begin{center}
\includegraphics[width=0.495\linewidth]{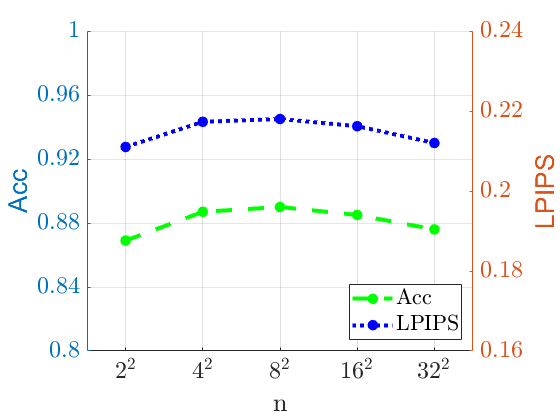}
\includegraphics[width=0.495\linewidth]{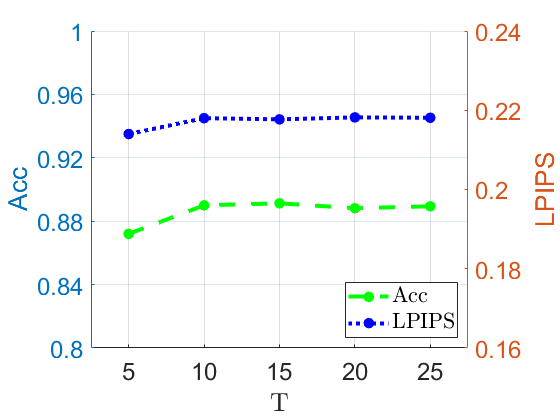}
\includegraphics[width=0.495\linewidth]{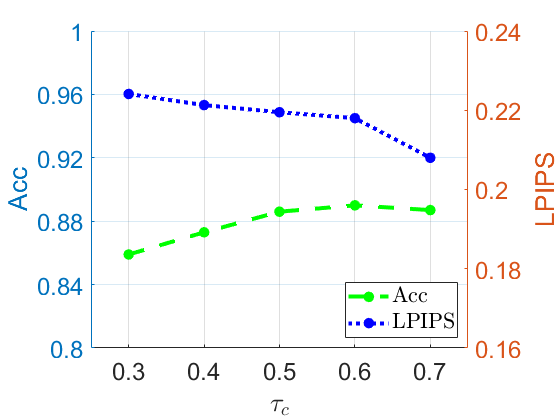}
\includegraphics[width=0.495\linewidth]{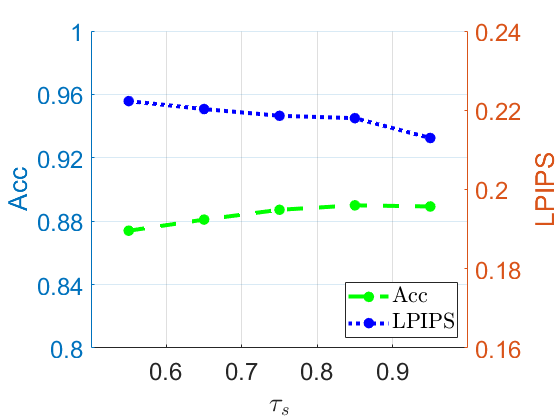}
\includegraphics[width=0.495\linewidth]{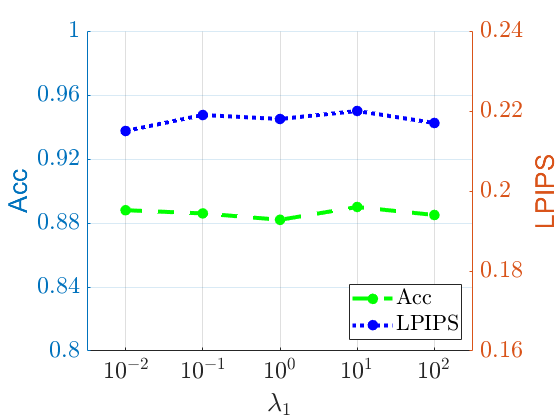}
\includegraphics[width=0.495\linewidth]{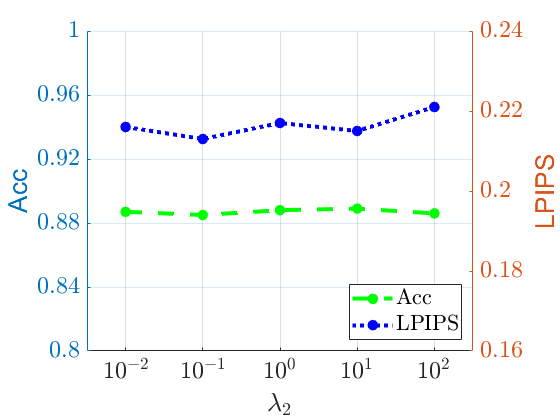}
\end{center}
\caption{Accuracy and LPIPS of our method when varying hyper-parameters $n$, $T$, $\tau_c$, $\tau_s$, $\lambda_1$, and $\lambda_2$.}
\label{fig:hyper_parameter}
\end{figure}

First, we explore spatial size $n=h\times w$ of the fused feature map $\bm{F}^m$, which decides the number of anchors.
We set $n=h\times w=8\times 8=64$ by default. Recall that the spatial size of extracted background feature map is $32\times 32$ and we use a $h\times w$ \emph{AdaptiveAvgPool} to downsample the background feature map to $h\times w$, as introduced in Section 4.3 in the main paper. We vary $n$ in $[2^2,4^2,8^2,16^2,32^2]$ and plot the results in Figure \ref{fig:hyper_parameter}, from which we observe that the performance first increases and then decreases. When the number of anchors is too small (\emph{e.g.}, $n=2^2$), the proposed method gets close to the generative method and the performance becomes worse, probably because of the difficulty in predicting plausible center offsets and foreground scales within a large scope. When the number of anchors is too large (\emph{e.g.}, $n=32^2$), each pixel-wise feature vector in the $h\times w$ background feature map may lack sufficient contextual information, leading to the performance drop. Therefore, we opt for $n=8^2$ by default. 

\begin{figure*}[t] 
\centering 
\includegraphics[width=0.93\textwidth]{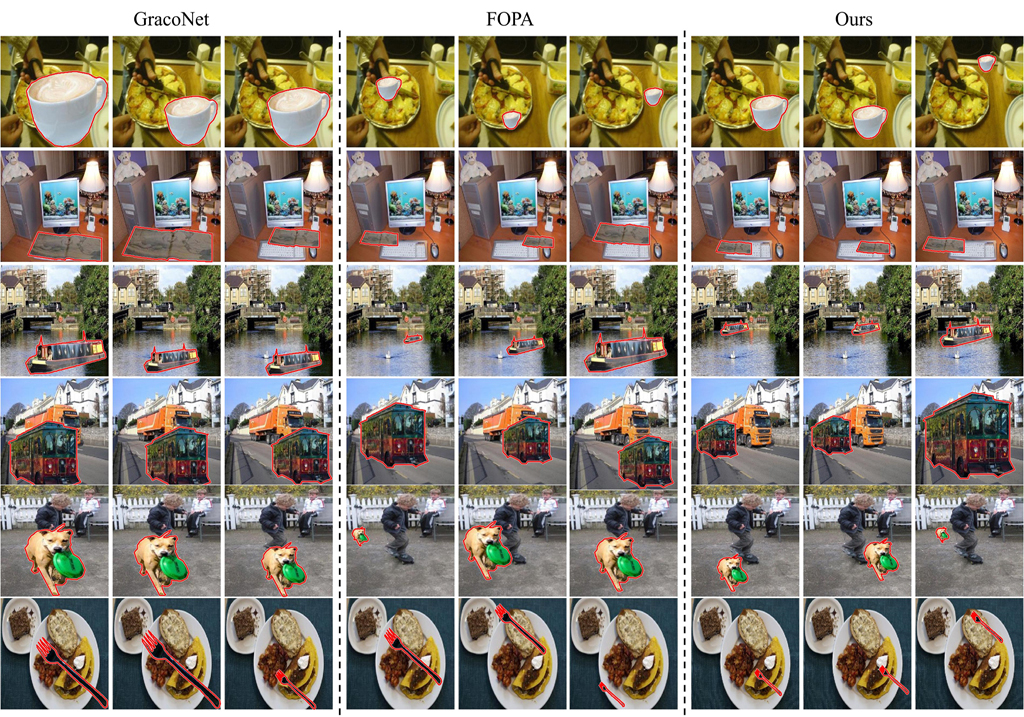}
\caption{For each foreground-background pair, we show three composite images generated by GracoNet~\cite{zhou2022learning}, FOPA~\cite{niu2022fast}, and our method.} 
\label{Fig:cmp_baselines_supp} 
\end{figure*}

We also study the effect of sampling times $T$ for each positive anchor. In OPA dataset \cite{liu2021opa}, we observe that one anchor in a test pair has at most $3$ positive placements, that is, $K_i \leqslant 3,\,\, \forall i$. With the constraint that $T  \geqslant K_i,\,\,\forall i$ (see Section 3.3 in the main paper), we vary $T$ in $[5, 10, 15, 20, 25]$ and plot the results in Figure \ref{fig:hyper_parameter}. We observe that as $T$ increases, the performance first gets better and then becomes stable. One possible explanation is that when $T$ is sufficiently large ($T  \geqslant 10$), our model pushes a wide range of generated composite images to be plausible and different from each other, which could benefit the plausibility and diversity of prediction results. Thus, we set $T=10$ by default. 

Then, we study the impact of two thresholds $\tau_c$, $\tau_s$ used in the self-training stage, which are set as $0.6$ and $0.85$ by default respectively.  
We vary $\tau_c$ in $[0.3, 0.4, 0.5, 0.6, 0.7]$ when fixing $\tau_s=0.85$, and vary $\tau_s$ in $[0.55, 0.65, 0.75, 0.85, 0.95]$ when fixing $\tau_c=0.6$. The results are plotted in Figure \ref{fig:hyper_parameter}. When $\tau_c$ is mall, accuracy is low because the added positive anchors could be very noisy and mislead the training process, but LPIPS is high probably due to the randomness introduced by noisy positive anchors. For small $\tau_s$, we have similar observation as $\tau_c$. In particular, when $\tau_s$ is mall, accuracy is low because the added positive placements could be very noisy and mislead the training process, but LPIPS is high probably due to the randomness introduced by noisy positive placements. 

Finally, we study the impact of $\lambda_1$ and $\lambda_2$ in Eqn. 7 in the main paper. We vary $\lambda_1$ (\emph{resp.}, $\lambda_2$) in the range of $[10^{-2}, 10^2]$ while fixing $\lambda_2$ (\emph{resp.}, $\lambda_1$) as the default value $5$ (\emph{resp.}, $10$).
The results are plotted in Figure \ref{fig:hyper_parameter}, from which we observe that our method is relatively robust to these two hyper-parameters.

\section{Details of User Study} \label{sec:user_study}
When evaluating the plausibility of generated composite images, we adopt user study as one of the evaluation metrics. Specifically, we
ask 50 voluntary participants to compare the composite images generated by TERSE~\cite{SyntheticTripathi2019}, PlaceNet~\cite{LearningObjPlaZhang2020}, GracoNet~\cite{zhou2022learning}, FOPA~\cite{niu2022fast}, and our method. 
For each test sample, every participant selects the method which generates the most plausible composite image. Then, we calculate the proportion that each method is selected as the best one among $50$ participants. Finally, we calculate the averaged proportion over all test samples as the user study score for each method, as reported in Table 1 in the main paper.

\begin{figure}[t]
\begin{center}
\includegraphics[width=0.99\linewidth]{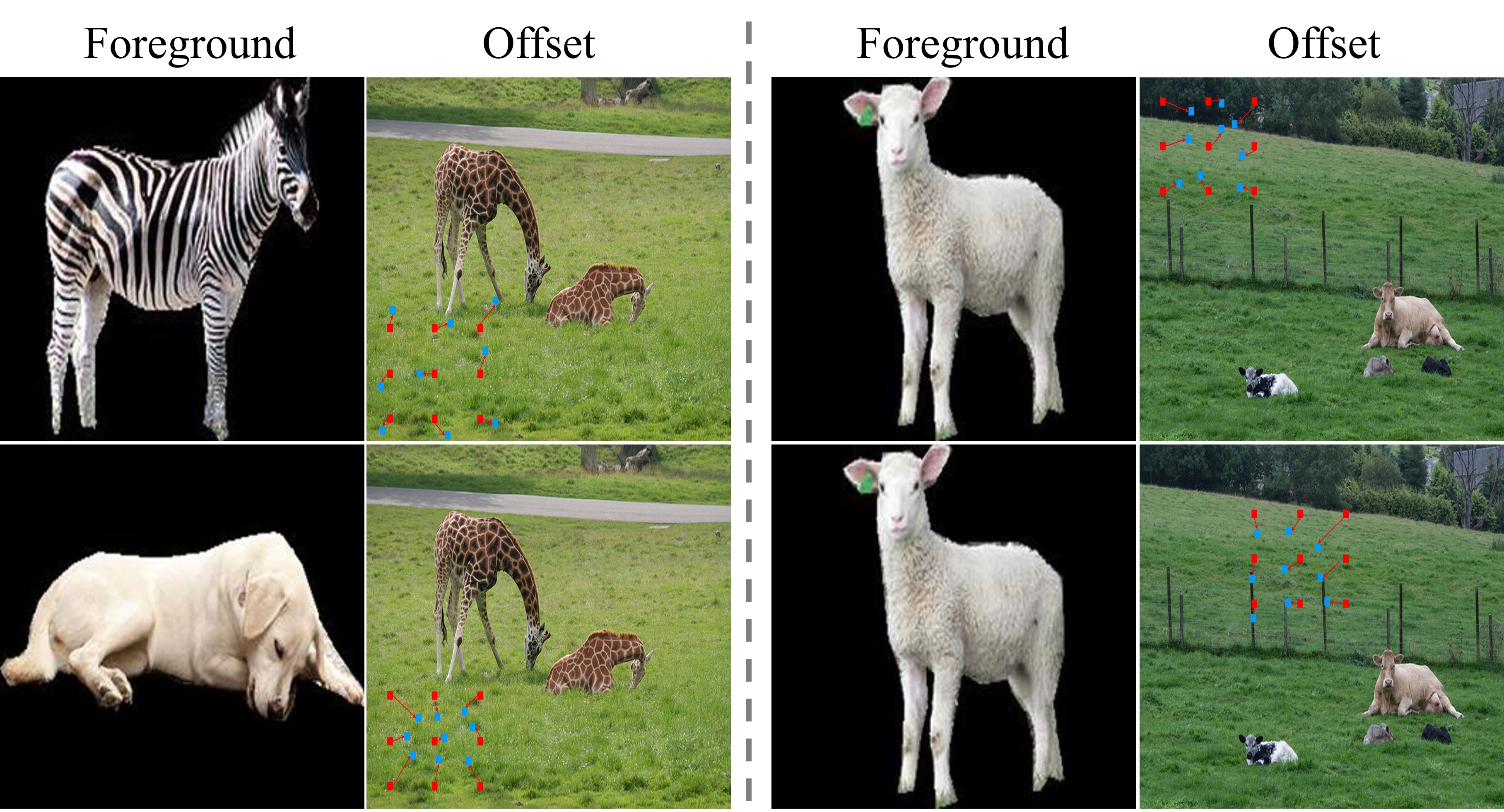}
\end{center}
\caption{Visualization of the convolution kernel offsets (red arrows) predicted by our dynamic fusion module (DFM).}
\label{fig:offset}
\end{figure}

\begin{figure*}[t]
\begin{center}
\includegraphics[width=0.99\linewidth]{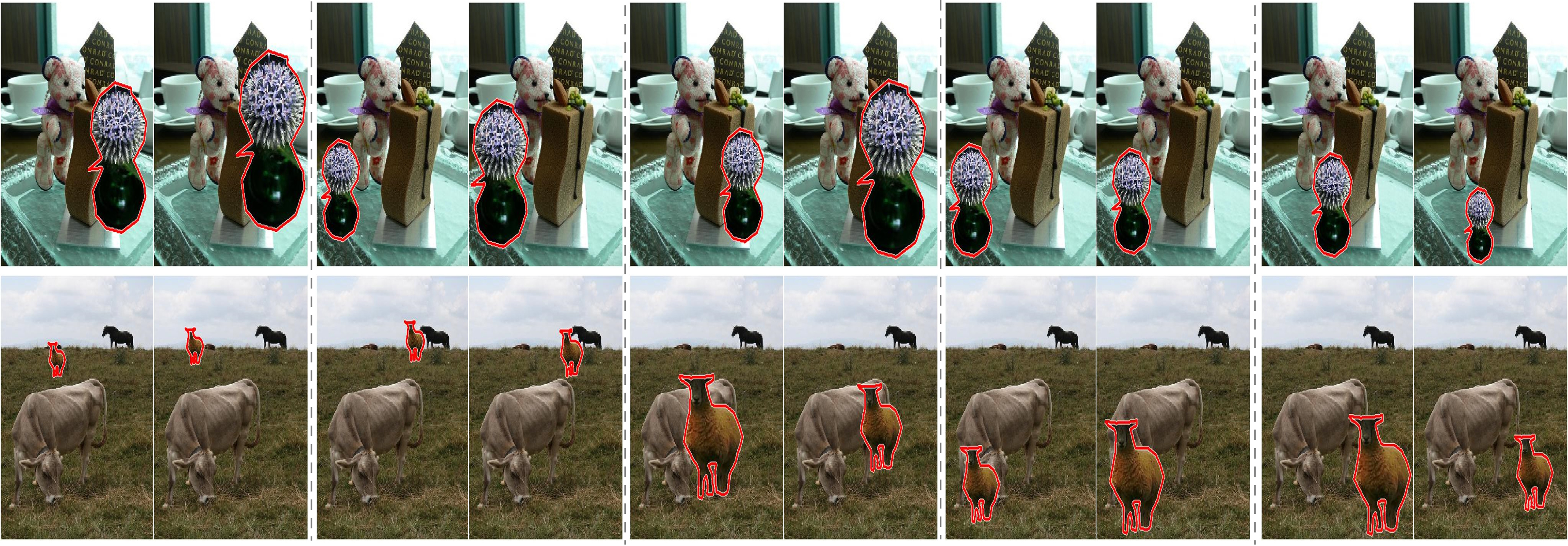}
\end{center}
\caption{For each foreground-background pair, we show 10 composite images generated by our method. Two composite images generated from the same positive anchor are grouped together. }
\label{fig:our_results}
\end{figure*}

\begin{figure}[t]
\begin{center}
\includegraphics[width=0.9\linewidth]{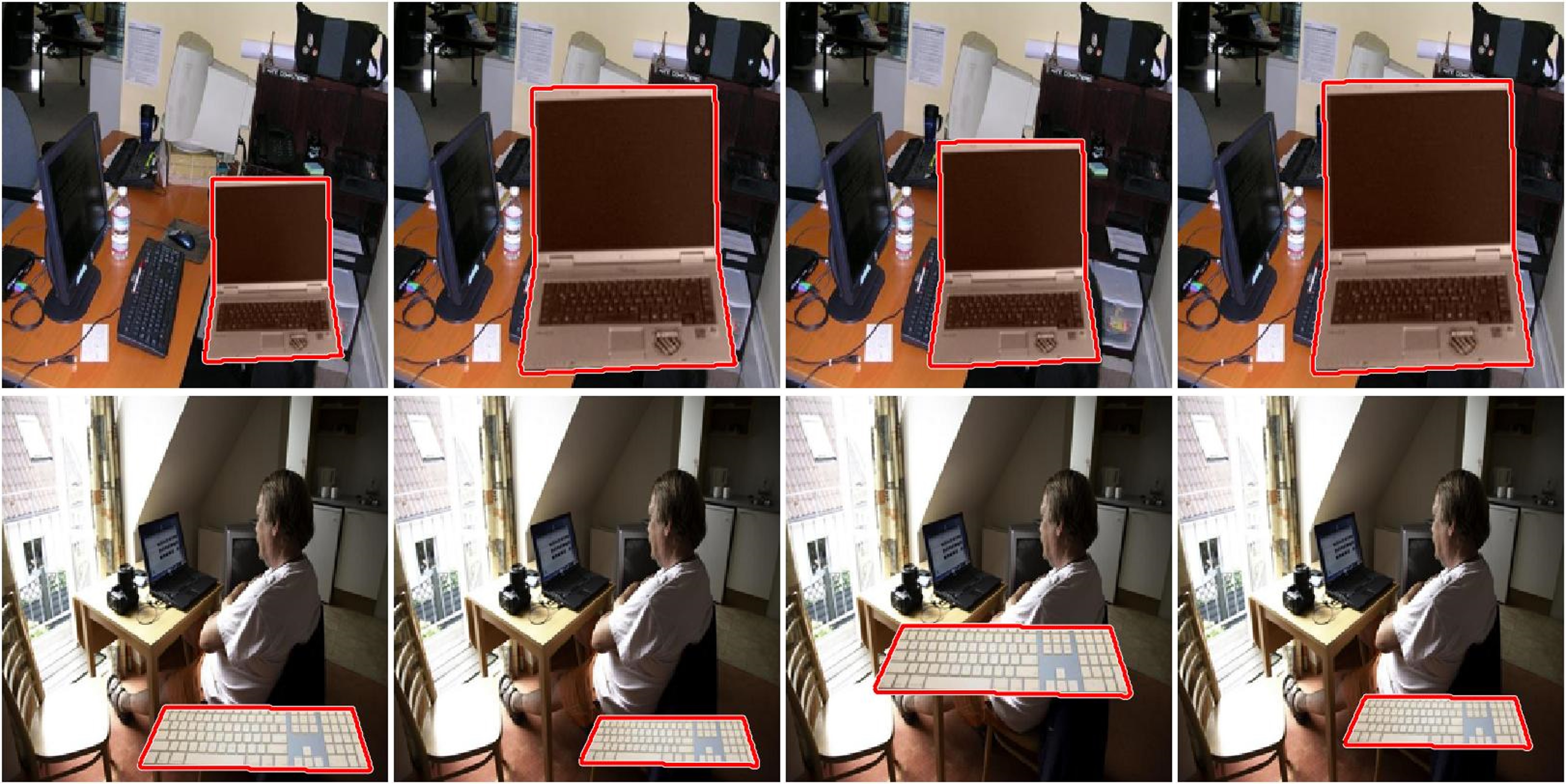}
\end{center}
\caption{Two example failure cases of our method.}
\label{fig:failure_cases}
\end{figure}

\section{More Visual Comparison with Baselines} \label{sec:visual_comparison}

We provide more visualization results to compare with the competitive baselines: GracoNet~\cite{zhou2022learning} and FOPA~\cite{niu2022fast}. In Figure \ref{Fig:cmp_baselines_supp}, we show three generated composite images for each method. 

We observe that GracoNet is prone to generate overlarge composite foreground (\emph{e.g.}, row 1, 2) and the location diversity is insufficient (\emph{e.g.}, row 3, 5). 
FOPA is able to produce more plausible and diverse composite images. Nevertheless, the foreground could be placed at unreasonable locations (\emph{e.g.}, the cup on the food in row 1) or with unreasonable scales (\emph{e.g.}, the bus in front of the truck is too small). 
The plausibility and diversity of composite images generated by our method is comparable or better than those of FOPA. Moreover, our method is much more efficient than FOPA (see Table 1 in the main paper). Therefore, our method can strike a good balance effectiveness and efficiency, compared with previous methods.

\section{Offset Visualization} \label{sec:offset_visualization}

Recall that we propose Dynamic Fusion Module (DFM) which dynamically aggregates different scales of contextual information based on foreground and background information. Specifically, we predict the offsets for convolution kernels based on the concatenation of background feature map and replicated foreground feature vector. In this section, we visualize the offsets predicted by our DFM in Figure \ref{fig:offset}, to show that they depend on foreground information and background context. In each example, given a pair of foreground and background, we choose an anchor location and visualize the predicted offsets for the convolution kernel centered at this location. 

By comparing two examples in the left column, we observe that at the same location on the same background, the physical size of foreground object affects the predicted offsets. When the foreground object has large physical size  (\emph{e.g.}, zebra \emph{v.s.} dog),  the predicted offsets are also large. Intuitively, for the large-scale foreground object, we need to speculate large-scale contextual information in the background (\emph{e.g.}, check the existence of unreasonable occlusion between the foreground object and background objects). 

By comparing two examples in the right column, we observe that for the same foreground and background, the location of kernel center also affects the predicted offsets. When there exists unreasonable occluder (\emph{e.g.}, fence) near the kernel center, one interesting observation is that the predicted offsets pay more attention to the occluders, probably because the detected unreasonable occluder is a strong evidence that this anchor is a negative anchor.

The above visualization results show that the predicted offsets depend on the foreground information and background context, which justifies our motivation to dynamically aggregate different scales of background contextual information.

\section{More Visualization Results of Our Method} \label{sec:our_visualization}

In this section, we provide more visualization results of our method. To better demonstrate the plausibility and diversity of our generated composite images, given a pair of foreground and background, we first randomly select five positive anchors (predicted rationality score larger than 0.5), and then generate two composite images for each positive anchor by randomly sampling the random vector twice. The generated composite images are shown in Figure \ref{fig:our_results}, and it can be seen that our method can usually generate diverse and plausible composite images considering different reasonable locations and scales.

\section{Failure Case}\label{sec:failure_cases}

Although our method can usually produce satisfactory results, our method may fail to predict reasonable scale and location in some challenging cases. For example, as shown in Figure \ref{fig:failure_cases}, when the background is very cluttered and there is only limited space (\emph{e.g.}, desk in row 2) for plausible placement, the results of our method are unsatisfactory.

{\small
\bibliographystyle{ieee_fullname}
\bibliography{egbib}
}